\documentclass[journal]{IEEEtran}

\usepackage{times}
\usepackage{epsfig}
\usepackage{graphicx}
\usepackage{amsmath}
\usepackage{amssymb}
\usepackage{subcaption}
\usepackage[noadjust]{cite}
\usepackage{xcolor}

\ifCLASSINFOpdf
\else
\fi
\begin{document}
%
% paper title
% Titles are generally capitalized except for words such as a, an, and, as,
% at, but, by, for, in, nor, of, on, or, the, to and up, which are usually
% not capitalized unless they are the first or last word of the title.
% Linebreaks \\ can be used within to get better formatting as desired.
% Do not put math or special symbols in the title.
\title{Exploiting Recurrent Neural Networks and Leap Motion Controller for Sign Language and Semaphoric Gesture Recognition}
%
%
% author names and IEEE memberships
% note positions of commas and nonbreaking spaces ( ~ ) LaTeX will not break
% a structure at a ~ so this keeps an author's name from being broken across
% two lines.
% use \thanks{} to gain access to the first footnote area
% a separate \thanks must be used for each paragraph as LaTeX2e's \thanks
% was not built to handle multiple paragraphs
%

\author{Danilo Avola,~\IEEEmembership{Member,~IEEE,}
        Marco Bernardi ,
        Luigi Cinque,~\IEEEmembership{Senior Member,~IEEE}
        Gian Luca Foresti,~\IEEEmembership{Senior Member,~IEEE}
        and Cristiano Massaroni% stops a space
}

\maketitle

% As a general rule, do not put math, special symbols or citations
% in the abstract or keywords.
\begin{abstract}
In human interactions, hands are a powerful way of expressing information that, in some cases, can be used as a valid substitute for voice, as it happens in Sign Language. Hand gesture recognition has always been an interesting topic in the areas of computer vision and multimedia. These gestures can be represented as sets of feature vectors that change over time. Recurrent Neural Networks (RNNs) are suited to analyse this type of sets thanks to their ability to model the long term contextual information of temporal sequences. In this paper, a RNN is trained by using as features the angles formed by the finger bones of human hands. The selected features, acquired by a Leap Motion Controller (LMC) sensor, have been chosen because the majority of human gestures produce joint movements that generate truly characteristic corners. A challenging subset composed by a large number of gestures defined by the American Sign Language (ASL) is used to test the proposed solution and the effectiveness of the selected angles. Moreover, the proposed method has been compared to other state of the art works on the SHREC dataset, thus demonstrating its superiority in hand gesture recognition accuracy.
\end{abstract}

% Note that keywords are not normally used for peerreview papers.

% For peer review papers, you can put extra information on the cover
% page as needed:
% \ifCLASSOPTIONpeerreview
% \begin{center} \bfseries EDICS Category: 3-BBND \end{center}
% \fi
%
% For peerreview papers, this IEEEtran command inserts a page break and
% creates the second title. It will be ignored for other modes.
\IEEEpeerreviewmaketitle

\section{Introduction}
Hands can express a lot of information thanks to the many gestures that their fingers can compose. There are different types of gestures depending on the kind of information that they intend to transmit. Based on the researches of Kendon~\cite{Kendon} and Quek et al.~\cite{Quek2002}, a taxonomy of possible gesture categories is proposed as follows:
\begin{itemize}
	\item \textbf{Deictic Gestures} are the gestures that involve pointing to establish the identity or spatial location of an object within the context of the application domain.
	\item \textbf{Manipulative Gestures} are usually performed by freehand movements to mimic manipulations of physical objects as in virtual reality interfaces.
	\item \textbf{Semaphoric Gestures} are particular gestures that define a set of symbols to communicate with machines.
	\item \textbf{Gesticulation} is one of the most natural forms of gesturing and it is commonly used in combination with conversational speech interfaces. These gestures are often unpredictable and difficult to analyse.
	\item \textbf{Language Gestures} are the gestures used for sign languages. They are performed by using a series of gestures that combine to form grammatical structures for conversational style interfaces. In case of finger spelling, these gesture can be considered as semaphoric.
\end{itemize}

Hand gesture recognition provides a means to decode the information expressed by the reported categories which are always more used to interact with innovative applications, such as interactive games~\cite{Hyung5711226,Rautaray6150485}, serious games~\cite{Avola2013490,Placidi2543345}, sign language recognition~\cite{Kim_6751299,Lu,Marin2016,Sohn_2012}, emotional expression identification~\cite{Barrientos_2002,Truong2016}, remote control in robotics~\cite{Calinon_2007,Goza_2004} or alternative computer interfaces~\cite{6883176,Pierce_2002,5720546,6470686}.
In general, the approaches used in hand gesture recognition can be divided into two main classes: 3D model-based~\cite{Cheng7208833} and appearance-based~\cite{Li_2013}. The first uses key elements of the body parts to acquire relevant 3D information, while the second one uses images or video sequences to acquire key features. 
In the past, several RGB cameras were necessary to obtain a 3D model of body parts, including hands. Recent works, supported by advanced devices, e.g., Microsoft Kinect~\cite{microsoft_Zhang}, LMC~\cite{s140203702}, and novel modelling algorithms based on depth map concept~\cite{Shotton_2011} have enabled the use of 3D models within everyday application domains.\\
In this paper, a language hand gesture recognition solution using 3D model-based approaches is presented. Specifically, the proposed method uses skeletal-based modelling, where a virtual representation of skeleton hands (or other parts of the body) is mapped to specific segments. This technique uses joint angle parameters along with segment lengths, instead of intensive processing of all 3D model parameters.
Then, it measures the variations over time of the skeleton joints whose spatial coordinates are acquired by a LMC. In particular, the angles formed by a specific subset of joints that involve distal, intermediate, and proximal phalanges for the index, middle, ring, and pinky, as well as the metacarpal for the thumb, can be considered highly discriminating to recognize many kind of hand gesture as confirmed by our tests. Moreover, these features have been selected as easy and quick to be extracted. Spatial information about the fingertips are also considered by the method in order to manage not articulated movements of the hands. 
In order to obtain a more accurate classifier, the information of the intra-finger angles and the spatial information of the palm of the hand are also considered.
During the design of the proposed method, two challenges were fixed: 
\begin{itemize}
	\item the search of a robust solution able to recognize also gestures that are similar to each other;
	\item the achievement of the highest accuracy level compared with works of the current literature.
\end{itemize}
These goals have been obtained by using a stack of RNNs~\cite{Graves6638947} with Long Short Term Memory (LSTM)~\cite{Sundermeyer_2015} architecture, a particular type of Deep Neural Network (DNN) where connections between its units form a directed cycle. The RNNs, unlike the common DNNs, can model long term contextual information of temporal sequences thus obtaining excellent results in the sound analysis and speak recognition, as reported in~\cite{Graves6707742}. The LSTM is an architecture where a RNN uses special units instead of common activation function. LSTM units help to propagate and preserve the error through time and layers. This aspect of the LSTMs allows the net to learn continuously over many time steps, thereby opening a channel to link causes and effects remotely. An architecture formed by two or more stacked LSTM RNNs is defined as Deep LSTM (DLSTM). Such an architecture allows to learn at different time scales over the input sequences~\cite{Hermans}. In the experiment section, we focused on a challenging subset of gestures defined by the ASL~\cite{Athitsos08theamerican}. This ASL is chosen because it is composed of a considerable number of types of gesture with several degrees of complexity.
The method proposed in this paper provides the following contributions:
\begin{itemize}
	\item the selection of a simple set of features based on the joint angles that are highly discriminative for the recognition of any kind of hand gesture, especially for language gestures;
	\item the capability of analyzing and recognizing large number of gestures in the field of language gesture classification. The study of static and dynamic gestures, belonging to the ASL, provides the prerequisites for achieving a wider recognition system for sign language;
	\item the DLSTM in combination with the skeleton extracted by the LMC has never been used before to recognize hand gestures. In addition, the LMC has been used in the design and collecting of a dataset, composed of a high number of gestures, that guarantees a high precision in the estimation of the positions of the joints of the hand~\cite{sensorLEAP}.
\end{itemize}
The rest of the paper is structured as follows. In Section~\ref{Related_work}, the state-of-the-art of the gesture recognition is provided. The proposed method is described in Section~\ref{Method}. Extensive experimental results are presented and discussed in Section~\ref{Experiments}. Finally, conclusions are drawn in Section~\ref{Conclusion}.
\begin{figure*}[t]
	\centering
	\includegraphics[width=\textwidth]{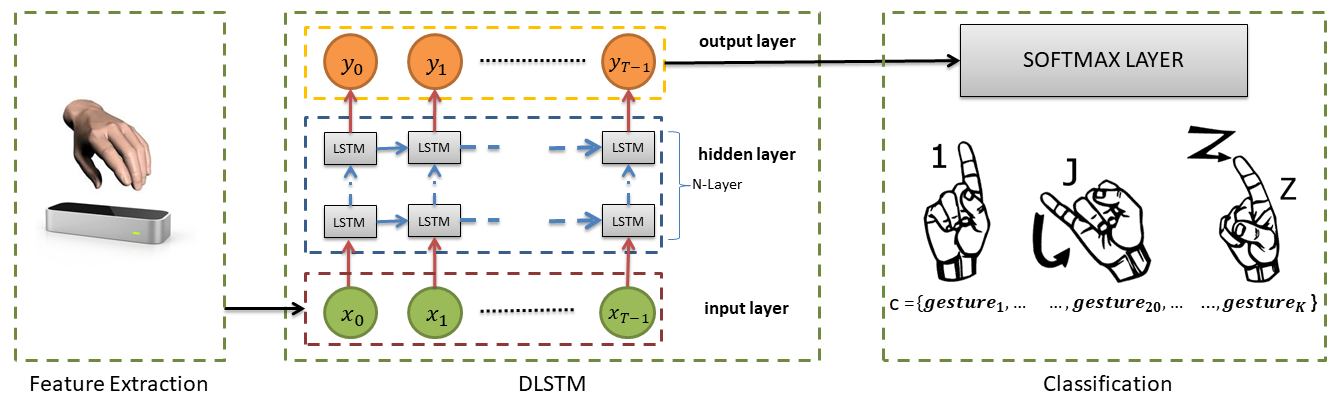}
	\caption{Logical architecture of the proposed method. The training phase is performed by a DLSTM with two stacked LSTM RNNs. Given a sequence of input vectors, the DLSTM returns an output vector for each time instant $t$, with $0 \leq t \leq T-1$, that contains the probabilities for each class. $K$ and $T$ are the different classes of the hand gestures and the maximum number of time instants in which a gesture is acquired, respectively.}
	\label{architecture}
\end{figure*}
\section{Related Work} 
\label{Related_work}
In the current literature, hand and body gesture recognition are based on a conventional scheme: the features are acquired from one or more sensors (such as Kinect \cite{ZHANG201529,6965622,DOMINIO2014101} and LMC \cite{Vikram2013,Lu}) and machine learning techniques (e.g., Support Vector Machine (SVM) \cite{Bar,Wang:2012}, Hidden Markov Models \cite{BorWang2006,Li_7410457} or Convolutional Neural Networks (CNNs) \cite{Neverova,Graves6707742}) are used to perform a classification phase.
A reference work is reported in \cite{Bar}, where SVM is used with Histogram of Oriented Gradients (HOG) as feature vectors. 
Wang et al. \cite{Wang:2012} and Suryanarayan et al.~\cite{Suryanarayan} used a SVM with volumetric shape descriptors. 
Using same classifier, Marin et al. \cite{Marin2016} applied a combination of features extracted by Kinect and LMC sensors.
Other interesting solutions are based on Hidden Markov Models (HMMs): Zun et al.~\cite{Zhu6298347} propose a robust hand tracking to recognize hand signed digit gestures.\\ 
Different well-known techniques are extended and customized to reach increasingly better results. An example is shown in ~\cite{Kim_6751299}, where a semi-Markov conditional model to perform finger-spelling gesture recognition on video sequences is reported. The Hidden Conditional Random Field (HCRF) method proposed in Wang et al.~\cite{BorWang2006} is instead used to recognize different human gestures. Lu et al.~\cite{Lu} use an extension of the HCRF to recognize dynamic hand gestures driven by depth data. Regarding the hand pose estimation, the solution proposed in Li et al.~\cite{Li_7410457} shows excellent results by applying a Randomized Decision Tree (RDT).\\
Another common solution is based on the use of Dynamic Time Warping (DTW). Although DTW does not belong to the class of machine learning techniques, it is often used in time series classification. In Vikram et al.~\cite{Vikram2013}, a DTW to support a handwriting recognition process based on the trajectory of the fingers extracted by a LMC is presented. In \cite{6975197}, the DTW with a novel error metric to match patterns, combined with a statistical classifier, is used to perform a tool to aid the study of basic music conducting gestures. In Sohn et al.~\cite{Sohn_2012}, a pattern matching method by the combination of a DTW and a simple K-Nearest Neighbor (K-NN) classifier is used.
Recently, the great performance of the deep neural networks has motivated the use of Convolutional Neural Networks (CNNs) in different application domains, including the gesture recognition as proposed in \cite{Neverova}. In addition, analysing the behaviour of these nets in other fields~\cite{Graves6638947,Sundermeyer_2015,Graves6707742}, we have understood that the RNNs can suitably support the classification of temporal data sequences. Based on these observations, the proposed method was designed starting from two works that achieve outstanding results in the current literature: the first, proposed by Du et al.~\cite{Du2015}, where an hierarchical RNN for skeleton action recognition is used, and the second, proposed by Graves et al.~\cite{Graves6707742}, using a Deep Bidirectional LSTM for the speech recognition.
\section{Method} 
\label{Method}
Let us consider, each hand gesture acquired by a user is represented by a set $X = \{x_0, x_1,..., x_{T-1}\}$ of feature vectors, where $T$ indicates the maximum number of time instants, inside a time interval $\Theta$, in which the features are extracted by a LMC. LMC is chosen as reference device for the acquisitions because it is optimized for the hands and the obtained skeleton model provides very accurate dynamic information about finger bones~\cite{leap}. A DLSTM is applied to model these sequences of data, where a time series of feature vectors (one vector for each time instant) is converted into a series of output probability vectors $Y = \{y_0, y_1,...,y_{T-1}\}$. Each $y_t \in Y$ indicates the class probability of the gesture carried out at time $t$, with $0 \leq t \leq T - 1$. Finally, the classification of the gestures is performed by a softmax layer~\cite{Bridle} using $K=|C|$ classes, where $C$ is the set of the considered gesture classes of the ASL. The logical architecture of the proposed method is shown in Fig.~\ref{architecture}. 
\subsection{Feature Extraction} 
\label{feat_extract}
\begin{figure}[!h]
	\centering
	\includegraphics[width=0.45\textwidth]{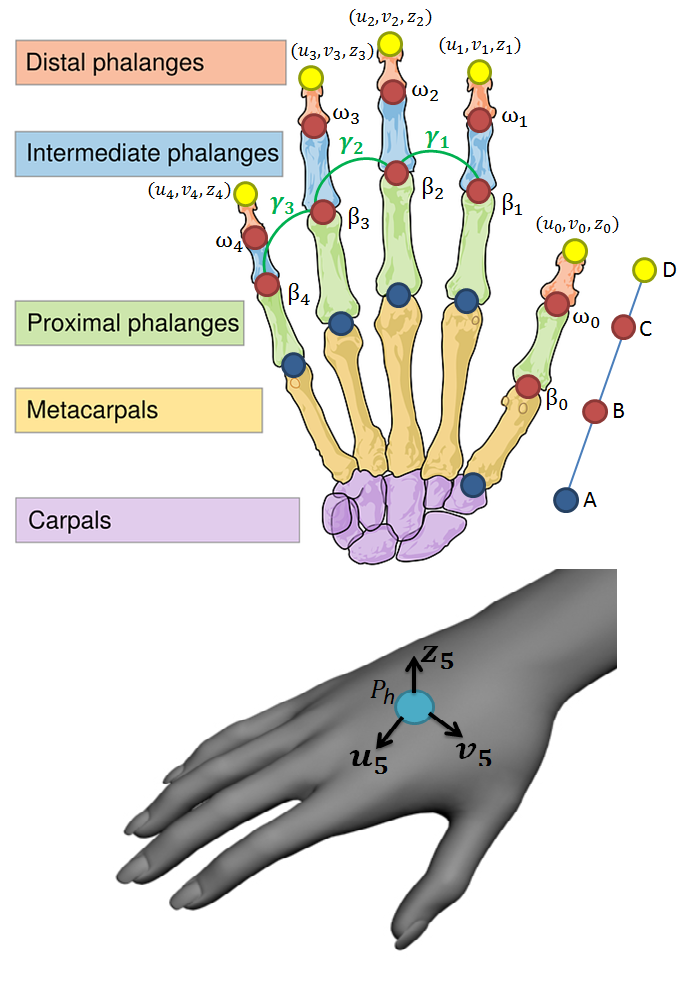}
	\caption{The features extracted from the hand: joint angles and fingertip positions. The yellow points indicate the fingertip positions on which the 3D displacements are computed. The red points indicate the joints on which the angles are computed.}
	\label{bones}
\end{figure}

Each gesture can be considered as the composition of different poses, where each pose is characterized by particular angles. Such a concept has already been applied in several works to recognize human actions, using the angles formed by the body joints \cite{6239231,6595915,6595916}.
So, each feature vector $x_t \in X$, with $0 \leq t \leq T-1$, is mainly composed by (Fig.~\ref{bones}):
\begin{itemize}
	\item the internal angles $\omega_1$, $\omega_2$, $\omega_3$, and $\omega_4$ of the joints between distal phalanges and intermediate phalanges. The internal angle $\omega_0$ considered for the thumb is computed between distal phalanx and proximal phalanx;
	\item the internal angles $\beta_1$, $\beta_2$, $\beta_3$, and $\beta_4$ of the joints between intermediate phalanges and proximal phalanges. The internal angle $\beta_0$ considered for the thumb is computed between proximal phalanx and metacarpal;
\end{itemize}

\begin{figure}[t]
	\centering
	\includegraphics[width=0.2\textwidth]{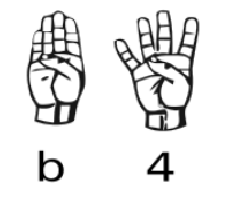}
	\caption{Example of static gestures differentiated by the intra finger angles $\gamma_1, \gamma_2$ and $\gamma_3$.}
	\label{Figure_02_2}
\end{figure}

Each finger can be seen as a set of segments, where $\overline{CD}$ is the distal phalanx, $\overline{BC}$ is the intermediate phalanx (with the exception of the thumb, where $\overline{BC}$ is the proximal phalanx), and $\overline{AB}$ is the proximal phalanx (with the exception of the thumb, where $\overline{AB}$ is the metacarpal). The angles are calculated as follows:
\begin{equation}
\omega_j = \arccos(\frac{ \overline{BC} \cdot \overline{CD} }{ |\overline{BC}||\overline{CD}|})
\end{equation}
\begin{equation}
\beta_j = \arccos(\frac{ \overline{AB} \cdot \overline{BC} }{ |\overline{AB}||\overline{BC}|})
\end{equation}

with $j = 0,..,4$.
Since the information provided by the angles is not sufficient to manage all types of existing gestures, especially dynamic gestures that perform movements in 3D space, additional information is used by considering the following features:
\begin{itemize}
	\item 3D displacements $u_5,v_5,z_5$ of the position of the central point $P_h$ of the palm of the hand. These features are considered to manage hand translation on the 3D space;
	\item 3D displacements $u_l,v_l,z_l$ of the fingertip positions, with $l = 0,..,4$. These features are considered to manage hand rotation in 3D space;
	\item the intra finger angles $\gamma_1, \gamma_2$ and $\gamma_3$, i.e., angles between two consecutive fingers. The fingers considered are the pointer finger, the middle finger, the ring finger and the pink finger. These features are used to handle special cases of static gestures that differ from each other only in intra finger angles, as shown in Fig.~\ref{Figure_02_2}.
\end{itemize}

All the listed features are independent by the reference.
Thus, the input vector assigned to the DLSTM at time $t$ is:
\begin{equation}
	x_t = \{\omega_0,...,\omega_4,\beta_0,...,\beta_4,u_0,v_0,z_0,...,u_5,v_5,z_5, \gamma_1,\gamma_2,\gamma_3\}
\end{equation}

\subsection{Sampling Process} 
\begin{figure}[!h]
	\centering
	\includegraphics[width=0.5\textwidth]{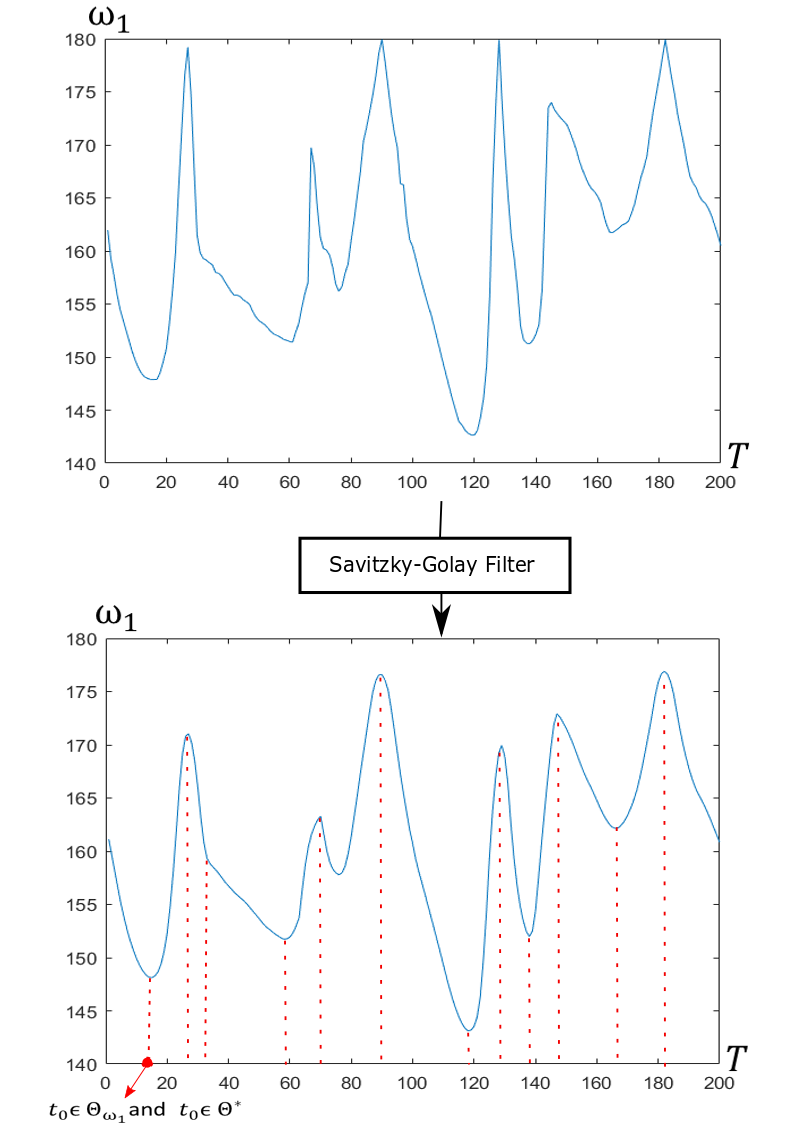}
	\caption{Sampling example for the feature $\omega_1$ on the "milk" gesture.}
	\label{sampling}
\end{figure}

As each person can perform the same gesture with different speeds, and as we want to analyze the sequences having all the same number $T$ of samples, we have implemented a sampling process able to select the most significant feature values within the entire time interval $\Theta$ of the gesture sequence. This means that data are acquired only in the most significant $T$ time instants, where an instant of time $t \in \Theta$ is defined as significant when the joint angles and the hand central point position $P_h$ vary substantially between $t$ and $t+1$ (as explained below). 
Let $f_{\omega_{i}}(t), f_{\beta_{i}}(t)$ and $f_{\gamma_{j}}(t)$, with $ 0 \leq i \leq 4$ and $1 \leq j \leq 3$, be the functions that represent the value of $\omega_i, \beta_i$ and $\gamma_j$ angles at time $t$, and let $f_{\phi_{(t)}}$ be the function that represents the value of $\phi$ (i.e., the displacement of the centre of the hand $P_h$ with respect to the previous position at time $t-1$) at time $t$.
For each function $f_g(t)$, with $g \in G$ and $G = \{\omega_i, \beta_i,\gamma_j, \phi\}$, the Savitzky-Golay filter \cite{Savitzky} is applied. The Savitzky-Golay filter is a digital filter able to smooth a set of digital data in order to increase the signal-to-noise ratio without greatly distorting the signal. Now, the significant variations on the considered features are identified through the relative maximum and minimum of each $f_g(t)$.
All the time instants $t$ associated with at least one relative minimum or relative maximum of a feature $g$ are used to create a new set $\Theta^{*}$, which represents a set of possible important time instants to sample.
In  Fig.~\ref{sampling}, an example of this sampling phase is shown, where the behaviour of the function $f_{\omega_1}(t)$ (the angle of the distal phalange of the index finger) for an instance of the gesture "milk" is considered. The signal in Fig.~\ref{sampling} is cleaned of any noise, caused by the acquisition device or tremors of the hand, by applying the Savitzky-Golay filter. Then, the maximum and minimum relative points are identified and sampled. In the example, only the procedure for the feature $\omega_1$ is shown, but this step is performed for each feature $g \in G$.
Now, depending on the cardinality of the set of the  sampled time instants, the following cases must be considered:
\begin{itemize}
	\item if $|\Theta^{*}|<T$, then the remaining $(|\Theta^{*}|-T)$ time instants to be sampled are randomly selected in $\Theta$;
	\item if $|\Theta^{*}|>T$, then, only some significant time instants are sampled for each $g$ feature. Let $\Theta_g$ be the set of the samples in $\Theta^{*}$ obtained from the relative maximum and minimum of the feature $g$ ($\Theta_g \subseteq \Theta^{*}$), we want to know the number of time instants $T_g$ that can be sampled for each $g$ such that $\sum_{g \in G} T_g=T$. Each $T_{g}$ is obtained thought the following proportion $|\Theta_g|:|\Theta^{*}|=T_g:T$. Then, from each $\Theta_g$ set, we will randomly take $T_g$ samples.
\end{itemize}

After the sampling step, each acquisition instance is composed by a sequence $\{x_0,..,x_{T-1} \}$ of feature vectors. The proposed sampling procedure is dynamically based on the value of the features.

\subsection{Deep Last Short Term Memory network}
\begin{figure}[!t]
	\centering
	\includegraphics[scale = 0.5]{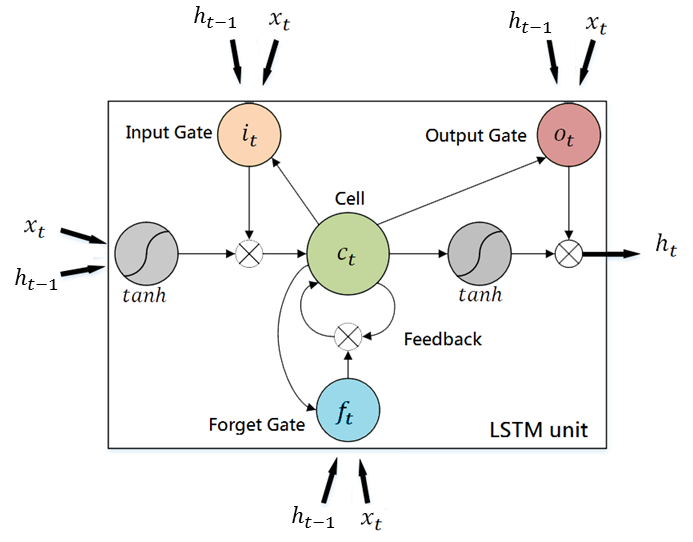}
	\caption{Example of a LSTM unit. The internal state is maintained with a recurrent connection. The input gate $i_t$ (orange) and the output gate $o_t$ (red) scale the input and output of the cell $c_t$, while the forget gate $f_t$ (azure) scales the internal state.}
	\label{LSTM}
\end{figure}

A fundamental component in the proposed work is the network used in the classification of the hand gesture. This network is based on multiple LSTMs, which unlike other types of NN, is able to efficiently analyze time sequences of data. Several factors, such as the error blowing up problem~\cite{Schaefer2008} and the vanishing gradient~\cite{Graves}, do not allow the use of common activation functions (e.g., tanh or sigmoid) to suitably train a network composed by multiple RNNs. This problem can be tackled with the LSTM units (Fig.~\ref{LSTM}). 

The LSTM can be seen as memory blocks that are one or more self-connected memory cells and three multiplicative units: the input, output and forget gates. These gates provide continuous analogues of write, read, and reset operations for the cells. Although LSTM allows to manage the problem of the vanishing gradient, the input time series often have a temporal hierarchy, with information that is spread out over multiple time scales which can not be adequately recognized by simple recurrent networks such as LSTMs.
For this reason, Deep RNNs or LSTMs were introduced. In fact, by constructing recurring networks formed over multiple layers, a higher abstraction on the input data is reached \cite{Hermans}. 
Increased input abstraction does not always bring benefits because the effectiveness of these networks depends on the task and the analysed input. In several works, such as \cite{Graves6707742,SakSB14,5373257}, it was observed empirically that Deep LSTMs work better than shallower ones on speech recognition. The audio signals, for example analysed in speech-to-text task, can be elaborated on more abstractions ranging from the entire pronounced phrase to the syllables of each word, and each abstraction can be captured in different time scales within the period considered.
Like in the case of audio sequences analysed in the speech recognition problem, gestures of the hand can be examined over multiple time scales. In fact, every gesture can be seen as the composition of so many small movements and sub-gesture of the hand, and its suitable for this type of network.
Based on these considerations, the LSTM stack-based solution have been experimented and then compared to the performance of a single-level network.
The first step is to define the activation functions of memory cell of the $LSTM_0$ (the first layer of the proposed neural network), as well as the input, output, and forget gates computed by using iteratively the following equations (from $t = 0~~to~~T-1$):
\begin{small}
	\begin{align}
		&i_{0,t} = \sigma( W_{xi}x_{t} + W_{hi}h_{0,t-1} + W_{ci}c_{0,t-1} + b_i )  \\
		&f_{0,t} = \sigma( W_{xf}x_{t} + W_{hf}h_{0,t-1} + W_{cf}c_{0,t-1} + b_f )  \\
		&c_{0,t}=f_t\odot c_{t-1}+i_{0,t}\odot \tanh(W_{xc}x_{t}+ W_{hc}h_{0,t-1}+b_c) \\
		&o_{0,t} = \sigma( W_{xo}x_{t} + W_{ho}h_{0,t-1} + W_{co}c_{0,t-1} + b_o )  \\
		&h_{0,t} = o_{0,t} \odot \tanh(c_{0,t})
	\end{align}
\end{small}
\hspace{-0.27cm}
where $i$, $f$, $o$, and $c$ denote the input gate, forget gate, output gate and cell activation vectors, respectively. These vectors have the same length of the hidden vector $h$. Instead, $W_{xi}$, $W_{xf}$, $W_{xo}$, and $W_{xc}$ are the weights of the input gate, forget gate, output gate and cell to the input. In addition, $W_{ic}$, $W_{fc}$, and $W_{oc}$ are the diagonal weights for peep-hole connections. Finally, the terms $b_i$, $b_f$, $b_c$, and $b_o$ indicates the input, forget, cell and output bias vectors, respectively. We have that $\sigma$ is the logistic sigmoid function and $\odot$ is the element-wise product of the vectors. Once the activation functions for the first level have been defined, the next step is to define the upper level activation functions.\\
\begin{figure}[t]
	\centering
	\includegraphics[width=0.47\textwidth]{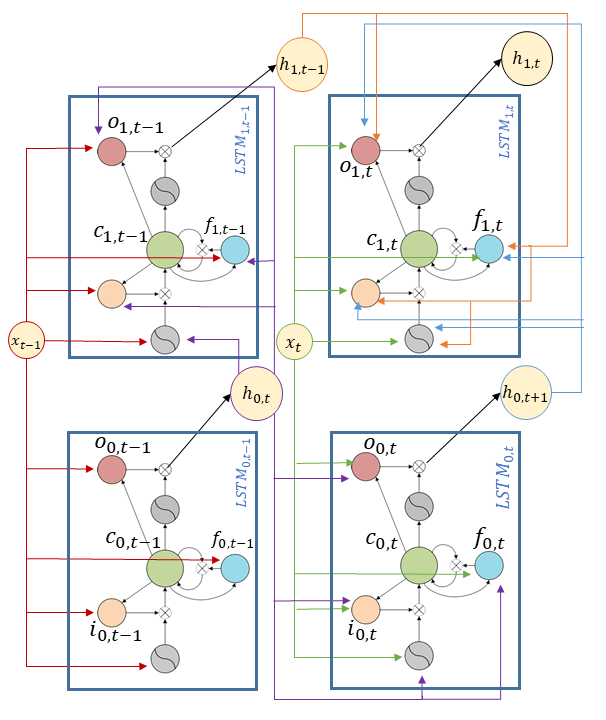}
	\caption{Example of connections between two stacked LSTM, where the first level is placed at the bottom of the image and is represented by the $LSTM_0$. For each level, the units that handle the inputs $x_{t-1}$ and $x_t$ are shown.}
	\label{DLSTM_Schema}
\end{figure}
DLSTMs are architectures obtained by stacking multiple LSTM layers where the output sequence $h_{l}$ of one layer $l$ forms the input sequence for the next layer $l+1$ (Fig.~\ref{DLSTM_Schema}). The memory cell of an $LSTM_l$ at time $t$, in addition to the classic $x_t$ and $h_{l,t-1}$ vectors, takes in input the $h_{l-1,t}$, i.e., the hidden state at time $t$ of the below $LSTM_{l-1}$. So, the activations of the memory cells of $LSTM_l$ of the network higher levels (i.e., $l>0$) are given by the following equations:
\begin{small}
	\begin{align}
		&
		\resizebox{0.9\hsize}{!}{$
		i_{l,t} = \sigma( W_{xi}x_{t} + W_{h_li}h_{l,t-1}+ W_{h_{l-1}i}h_{l-1,t} + W_{ci}c_{l,t-1} + b_i )
		$}  \\
		&
		\resizebox{0.9\hsize}{!}{$
		f_{l,t} = \sigma( W_{xf}x_{t} + W_{h_lf}h_{l,t-1}+ W_{h_{l-1}f}h_{l-1,t} + W_{cf}c_{l,t-1} + b_f ) 
		$} \\
		&
		\resizebox{0.9\hsize}{!}{$
		c_{l,t}=f_t\odot c_{t-1}+i_{l,t}\odot \tanh(W_{xc}x_{t}+ W_{h_{l}c}h_{l,t-1}+ W_{h_{l-1}c}h_{l-1,t} + b_c)
		$} \\
		&
		\resizebox{0.9\hsize}{!}{$
		o_{l,t} = \sigma( W_{xo}x_{t} + W_{h_lo}h_{l,t-1}+ W_{h_{l-1}o}h_{l-1,t} + W_{co}c_{l,t-1} + b_o ) $} \\
		&
		h_{l,t} = o_{l,t} \odot \tanh(c_{l,t})
	\end{align}
\end{small}
The output of the DLSTM network, at time $t$, with $N$-layers, is given by:
\begin{equation}
	y_t = W_{h_{N-1},t} h_{N-1,t} + b_y 
\end{equation}
where $b_y$ is a bias vector, $h_{N-1,t}$ is the hidden vector of the last layer and $W_{h_{N-1},t}$ is the weight from the hidden layer $h_{N-1,t}$ to output layer. The output $y_t$ defines a probability distribution over the $K$ possible gesture classes, where $y_t^{k}$ (the $k^{th}$ element of $y_t$) is the estimated probability of a specific class $C_k$ at time $t$ for the acquired gesture $X$. Finally, all results $y_t$ are collected and normalized into the softmax layer, through the following equations:
\begin{equation}
	\hat{y} = \sum_{t=0}^{T-1} y_t 
\end{equation}
\begin{equation}
	\tilde{y}^k =	p(C_k|X) = \frac{e^{\hat{y}^k}}{\sum_{ q = 0}^{K-1} e^{\hat{y}^q}}
\end{equation}
for each $k$, with $1 \leq k \leq K$. The classification of gesture $X$ will be given by the highest probability contained in $\tilde{y}$.

\subsection{Network Training} 
\label{NetworkTraining}
Given a dataset $D$ composed of $M$ train gesture sequences, the goal is to minimize the following maximun-likelihood loss function:
\begin{equation}
	\mathcal{L}(D) = - \sum_{m=0}^{M-1}\ln \sum_{k=0}^{K-1} \delta(k,\tau)p(C_k|D_m)
\end{equation}
where $D_m$, $ 0 \leq m \leq M$, is an input sequence of the training dataset $D$,   $\tau$ is the ground-truth label of $D_m$ and $\delta(\bullet, \bullet)$ is the Kronecker delta or delta function. This formulation is referred to the cross-entropy error proposed in~\cite{Bishop_2006}. The Back-Propagation Through Time (BPTT) algorithm~\cite{Graves} is used to obtain the objective function derived with respect to all the weights and to compute the minimization based on the stochastic gradient descent.

\begin{figure}[!h]
	\centering
	\begin{subfigure}{0.5\textwidth}
		\includegraphics[width=\textwidth]{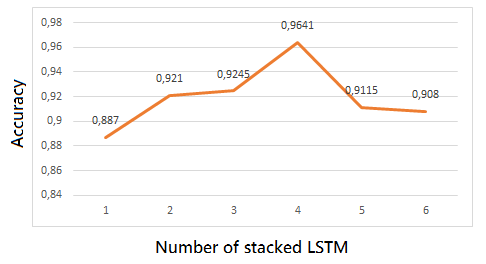}
		\caption{}   
		\label{level_accurancy_a}
	\end{subfigure}
	\begin{subfigure}{0.5\textwidth}
		\includegraphics[width=\textwidth]{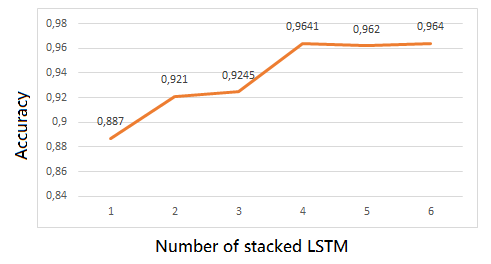}
		\caption{}   
		\label{level_accurancy_b}
	\end{subfigure}
	\caption{Accuracy results on the proposed dataset by varying the number of stacked LSTMs in the network architecture. (a) Accuracy results using 800 epochs for each considered architecture and (b) accuracy results using 800 epochs for 1-LSTM, 2-LSTM, 3-LSTM and 4-LSTM; for the 5-LSTM and 6-LSTM are used 1600 and 1800 epochs, respectively. The x-axis indicates the number of the stacked LSTMs, while the y-axis indicates the accuracy values.}
	\label{level_accurancy}
\end{figure}

\begin{figure*}
	\centering
	\includegraphics[scale = 0.50]{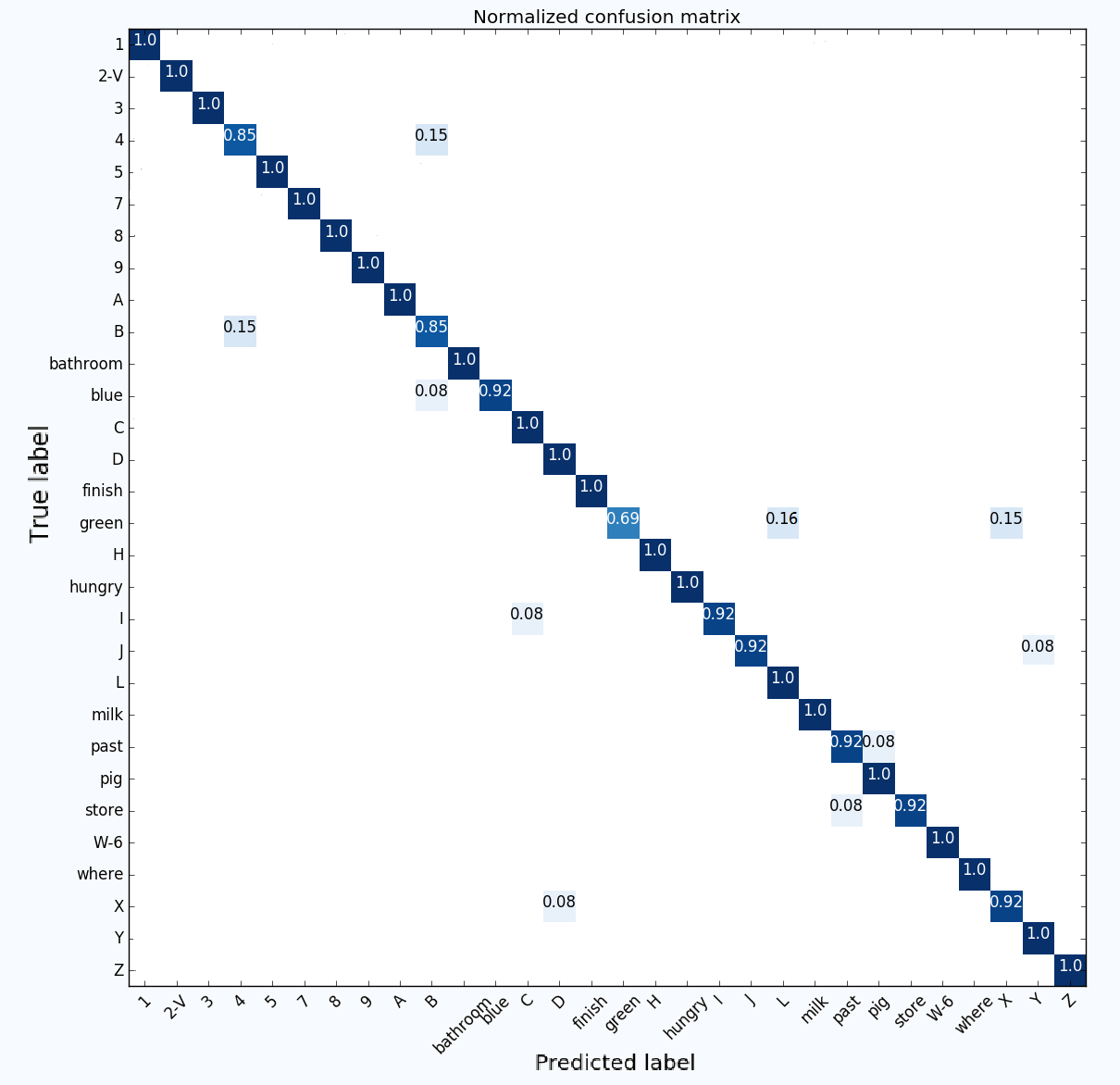}
	\caption{The confusion matrix related to the proposed gesture dataset. The overall accuracy is 96.4102\%.}
	\label{confusion_matrix}
\end{figure*}

\section{Experimental Results}
\label{Experiments}
This section describes the experiments performed to evaluate the behavior of the proposed approach. All the experiments were performed by using a LMC on an Intel i5 3.2GHz, 16GB RAM, with a GeForce GTX 1050ti graphics card. The DLSTM network and the BPTT algorithm, used to compute the minimization based on the stochastic gradient descent, are implemented by using the Keras\footnote{https://keras.io/} framework. The main purposes of the experiments were the assessment of the joint angles as salient features for the recognition of the hand gestures, the overall robustness of the proposed method and its higher accuracy compared with the current literature. The experiments were performed by using a challenging subset of gestures defined by the ASL and described in Section~\ref{Dataset}. A discussion of the obtained results is presented in Section~\ref{Result}, and a comparison of the proposed method with key works of the current state-of-the-art is reported in Section~\ref{compstate}.

\subsection{Dataset} 
\label{Dataset}

Currently, there is no public ASL dataset with a large number of classes and with information on the hand joint. Based on these reasons, we acquired the data of 30 gestures to create a new dataset.
This dataset consists of 12 dynamic gestures and 18 static gestures taken by the ASL. These additional gestures are chosen to represent much of the variations in joint angles and finger positions that occur when the hand perform a gesture. The static gestures are: 1, 2-V, 3, 4, 5, 6-W, 7, 8, 9, A, B, C, D, H, I, L, X and Y. indeed, the dynamic gesture are: bathroom, blue, finish, green, hungry, milk, past, pig, store and where. The dataset is composed of 1200 hand gesture sequences, coming from 20 different people. Each gesture was collected by 15 males and 5 females, aged 20 to 28 years. Each person performed the $K$ hand gestures twice, once for each hand. $K=30$ different gestures are considered.
The sequences from 14 people are used to create the train set while sequences of the remaining 7 people were used to form the test set. So, the 7 people used in the tests are never taken into consideration during the training phase.
As previously described in Section~\ref{feat_extract}, each sequence is acquired according to a sampling process, with $T = 200$ and $\Theta = 5s$.

\begin{table*}
	\centering
	\resizebox{\textwidth}{!}{
		\begin{tabular}{lc c c c c c}
			 & $\omega_i$ & $\beta_i$ & $\gamma_{j}$ &  $u_{w},v_{w},z_{w}$ & $\omega_i , \beta_i$ & $\omega_i , \gamma_{j} , \beta_i$  \\ [0.5ex]
			\hline                                          
			accurancy\% & 62.70\% & 68.1204 \% & 46.67\% & 56.92\% & 79.74\% & 85.13\%\\ [0.5ex]
			\hline
		\end{tabular}
	}
	\caption{Accurancy of the proposed solution obtained on datasets of 30 by varying the features, where $0 \leq i \leq 4$,  $0 \leq j \leq 3$ and $0 \leq w \leq 5$.} 
	\label{resu_tab_angle} 
\end{table*}

\begin{table}
	\centering
	\resizebox{\columnwidth}{!}{
		\begin{tabular}{lc c c}
			Accurancy & Precision & Recall & F1-Score       \\ [0.5ex]
			\hline                                          
			96.4102\% & 96.6434\% & 96.4102\% & 96.3717\%  \\ [0.5ex]
			\hline
		\end{tabular}
	}
	\caption{Performance of the proposed solution on dataset of 30 gestures using Precision, Recall, and F1-Score metrics.} 
	\label{resu_tab} 
\end{table}

\subsection{Selection of the Optimal Number of Stacked LSTM}\label{lstm_number_section}
Several tests have been conducted to chose the optimal number of stacked LSTMs to use in the proposed architecture. The hidden units per LSTM are $200$, i.e., the hidden units are equal to the number of input time instances considered for each gesture.
In Fig.~\ref{level_accurancy_a}, it is shown as an architecture composed by 4 levels gives the best accuracy results using 800 epochs.
In fact, although several levels of LSTM allow to analyze complex time sequences by dividing them into multiple time scales, the 5-LSTM and the 6-LSTM require more epochs to be trained. Increasing the number of epochs needed to train the 5-LSTM and 6-LSTM architectures (i.e., 1600 epochs for the 5-LSTM and 1800 for the 6-LSTM), the Fig.~\ref{level_accurancy_b} shows how their results improves. We can notice how greater abstraction on input does not provide substantial benefits from a certain number of levels, and the accuracy gained by the network begins to converge to a precise value. In conclusion, 4 levels are appropriated for the proposed network and represent a good compromise between training times and system accuracy. The choice of the learnig rate influences the speed of the convergence of the cost function. 
If the learning rate is too small, the convergence is obtained slowly, while if the learning rate is too large, the cost function may not decrease in every iteration and therefore it could not converge. 
In the proposed method, the learning rate is set to $0.0001$ through large empirical tests.

\begin{figure}[!h]
	\centering
	\begin{subfigure}{0.24\textwidth}
		\includegraphics[width=\textwidth]{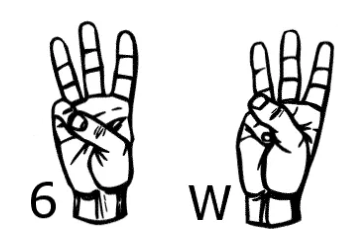}
		\caption{6 and W gestures.}   
		\label{6-W}
	\end{subfigure}
	\begin{subfigure}{0.24\textwidth}
		\includegraphics[width=\textwidth]{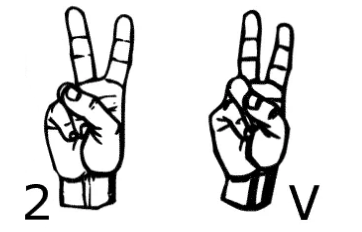}
		\caption{2 and V gestures.}   
		\label{2-V}
	\end{subfigure}
	\caption{Pairs of gestures joined into a common class.}
	\label{exception_class}
\end{figure}

\subsection{Feature Effectiveness Analysis}

In order to verify the effectiveness of the features in classifying the set of gestures taken by the ASL dataset, some tests have been carried out. The tests carry out the training of the network and the subsequent classification (using two different sets to train and test the network, respectively) of the ASL gestures, where the number of stacked LSTM is 4 (as explained in Section \ref{lstm_number_section}), using only subsets of $x_t$ as features. The results in Table~\ref{resu_tab_angle} shown as the combination of $\omega_i$ and $\beta_i$ features is able to discern alone an high number of gestures of the ASL vocabulary and it reaches better classification results with respect to separate use of these two features. Although the single $\gamma_j$ feature does not offer good performance, it greatly improves classification when used with $\omega_i$ and $\beta_i$. Instead, the combination of features that relate to the movements of the hand ($u_{w},v_{w},z_{w}$) are unable by themselves to classify the gesture of the hand but, if combined with the features of the angles, allow the method to achieve high performance (as discussed to follow).

\subsection{Hand gesture recognition on the ASL dataset}\label{Result}
To evaluate the method, we have used very popular metrics. First of all, we have used the accuracy as main metric. In addition, we have also introduced precision, recall and f1-score metrics can be considered a de facto standard to measure the quality of this class of algorithms~\cite{Sokolova_2009}. the obtained results are presented in Table~\ref{resu_tab}.
To better analyse the proposed approach and according to the tests performed to recognize the different hand gestures, the confusion matrix is computed (Fig.~\ref{confusion_matrix}). Each column of the matrix represents the instances in a predicted gesture, instead each row represents the instances in a current gesture. The main diagonal of the matrix represents the instances correctly classified by the DLSTM. The elements below the diagonal represent the false positives, i.e. the gestures that are incorrectly classified within a class of interest. The elements above the diagonal are the false negatives, i.e., the gestures incorrectly classified as not belonging to a class of interest. The distinction of some gestures is very hard, since they are very similar to other gestures in the dataset. Despite this, the proposed method does not suffer of ambiguity issues. The only exceptions are given by the gestures 6 with W (Fig.~\ref{6-W}) and 2 with V (Fig.~\ref{2-V}). The variations in their joint angles are minimal and difficult to see even to the human eye: moreover, the LMC device fails to capture these variations. For this reason, these gestures have been gropued in the same class. Tests performed without grouping these classes achieved 91.5178\% of accuracy. In fact, the decrease in accuracy is caused precisely by some incorrect classifications regarding the 2,6,v and W classes.

\begin{figure}[!h]
	\centering
	\begin{subfigure}{0.5\textwidth}
		\includegraphics[width=\textwidth]{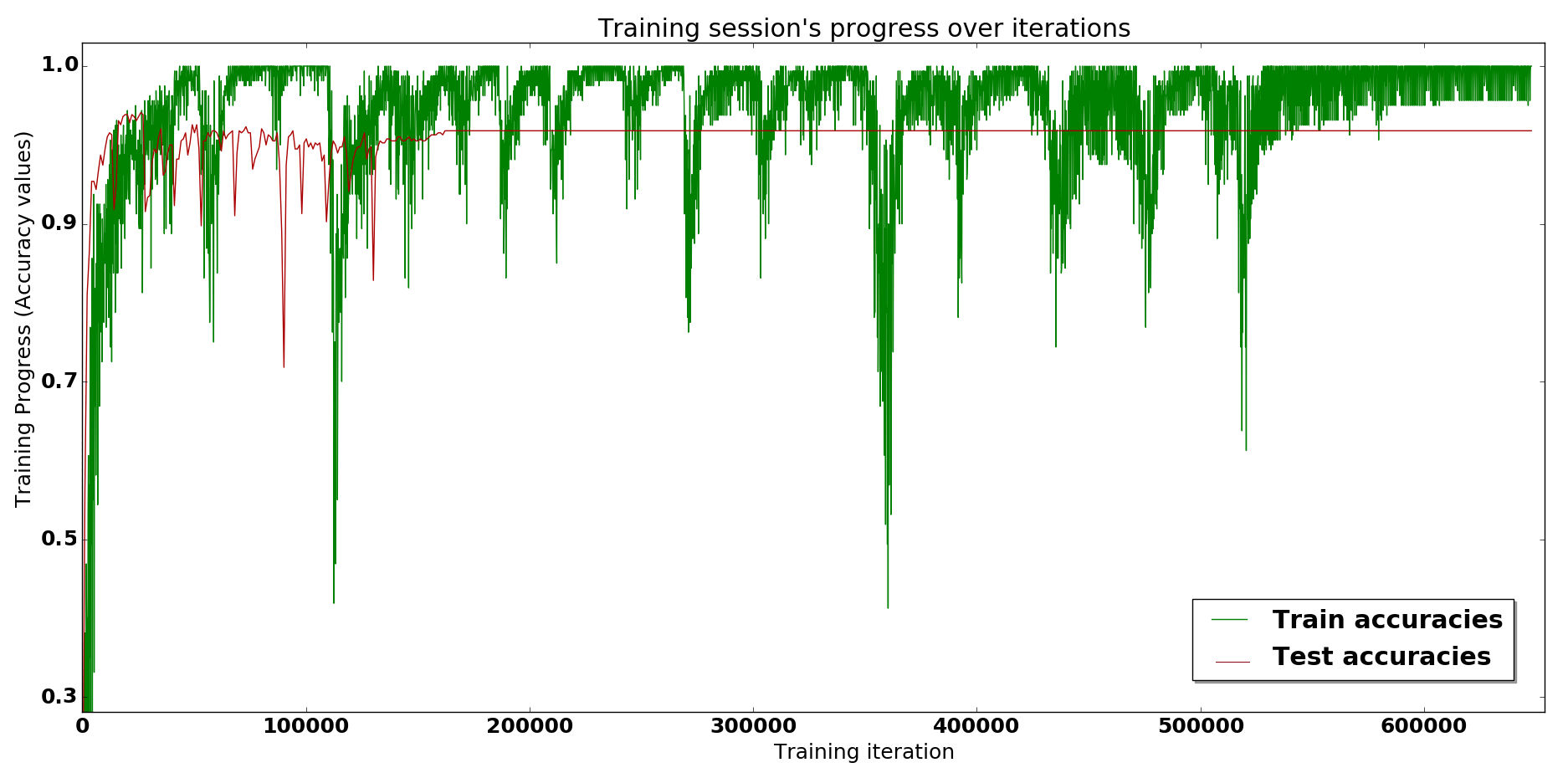}
		\caption{}   
		\label{accuracy_plot}
	\end{subfigure}
	\begin{subfigure}{0.5\textwidth}
		\includegraphics[width=\textwidth]{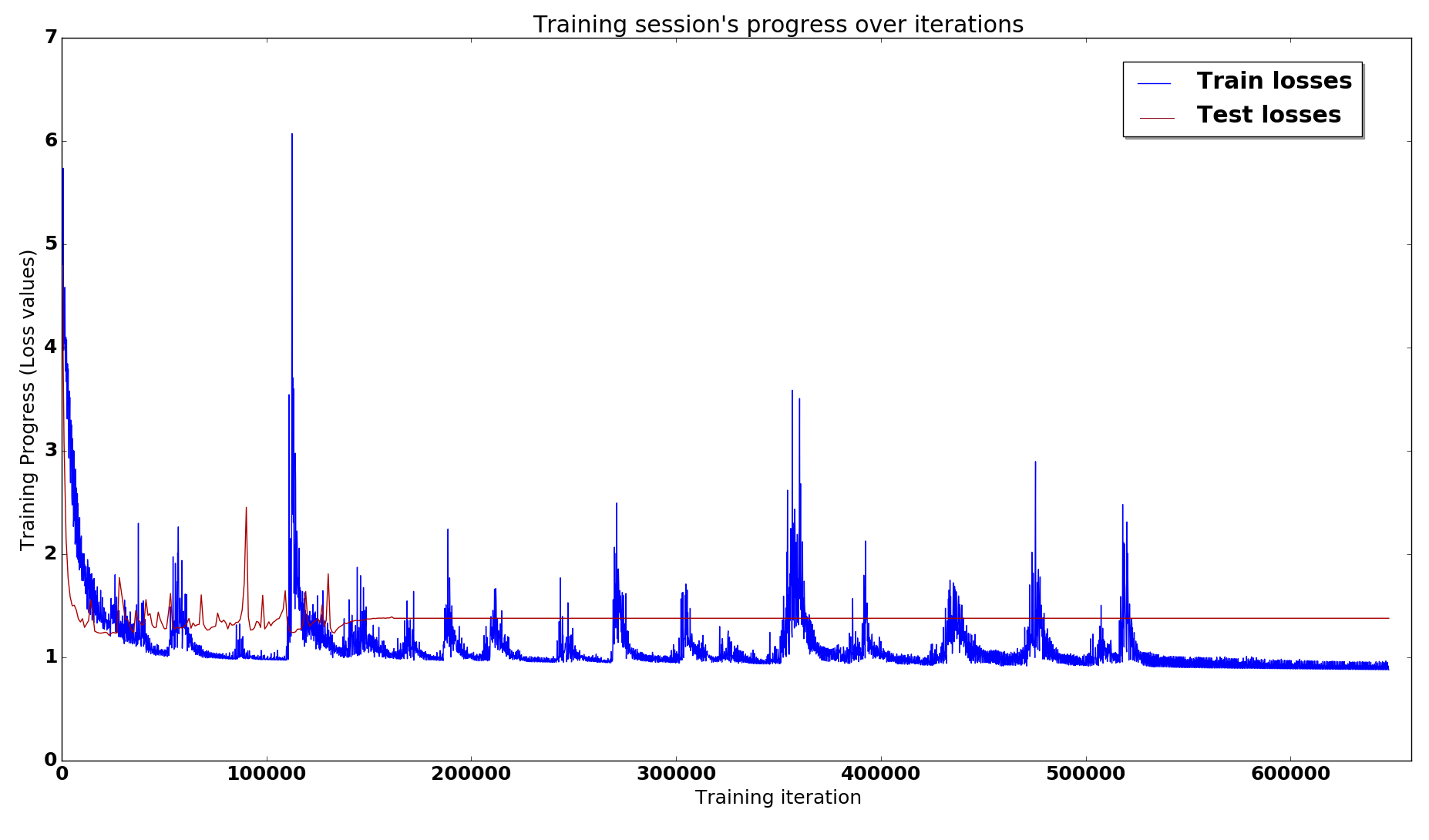}
		\caption{}   
		\label{loss_plot}
	\end{subfigure}
	\caption{Train/Val Curves: The progress of training and testing over iteration based on (a) the accurancy and (b) the loss. After 100000 iterations, the test accuracy curve converges. The x-axis represents the progress of training/validation stage and the y-axis represents the number of training iterations.}
	\label{plot}
\end{figure}

In Fig.~\ref{plot}, the Train/Test plots are shown. The first plot (Fig.~\ref{accuracy_plot}) shows the Train/Test accuracy over the iterations, instead the second plot (Fig.~\ref{loss_plot}) contains the loss curves that represent the sum of the errors provided for each training or test instance. In this work, the loss curves are calculated as maximum-likelihood loss function, described in Section~\ref{NetworkTraining}. Instead, the curves of accuracy represent the training or validation instances correctly recognized. After a certain number of iterations ($\sim 125000$), the test accuracy curve converges.
\begin{figure}[!h]
	\centering
	\begin{subfigure}{0.5\textwidth}
		\includegraphics[width=\textwidth]{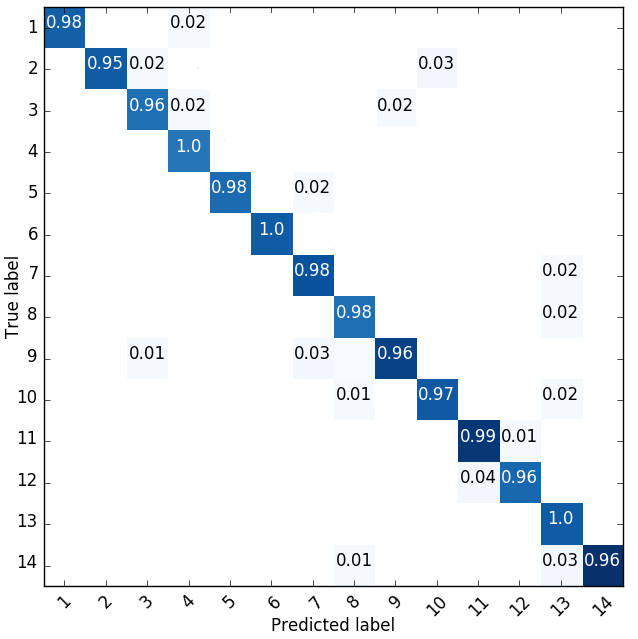}
		\caption{}   
		\label{SHREC14}
	\end{subfigure}
	\begin{subfigure}{0.5\textwidth}
		\includegraphics[width=\textwidth]{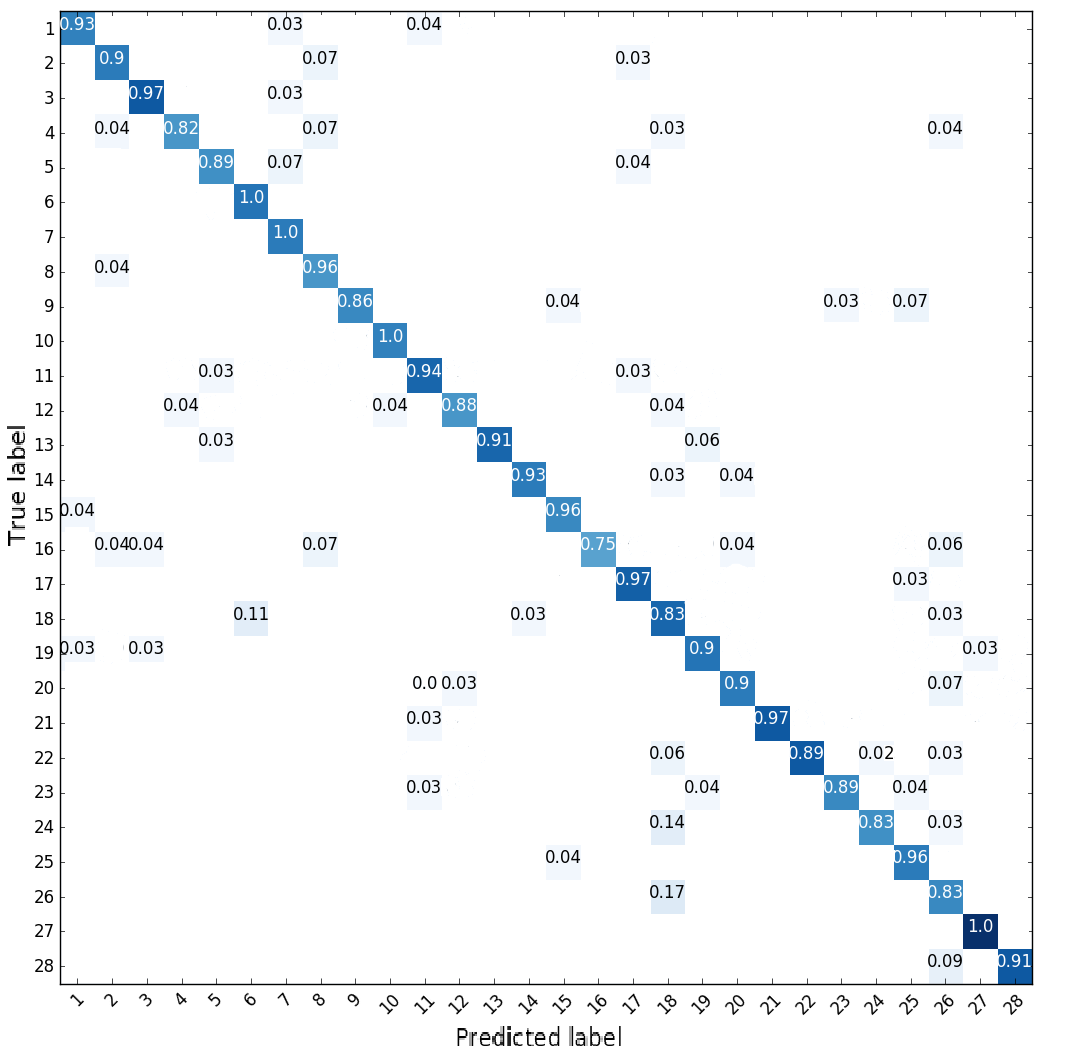}
		\caption{}   
		\label{SHREC28}
	\end{subfigure}
	\caption{Confusion matrices obtained on the SHREC dataset (a) with 14 gestures (b) and with 28 gestures.}
	\label{comparison_confusion_matrix}
\end{figure}
\begin{table*}
	\centering
	\resizebox{\textwidth}{!}{%
		\begin{tabular}{l|c|c}
			Features & Accuracy 14 Gestures & Accuracy 28 Gestures \\
			\hline
			Proposed Method & \textbf{97.62\%}  & \textbf{91.43\%} \\
			Skeleton-based Dynamic hand gesture recognition \cite{7789643} & 88.24\% & 81.90\% \\
			Key frames with convolutional neural network \cite{desmedt} &82.90\% &71.90\%\\
			Joint Angles Similarities and HOG2 for Action Recognition \cite{6595915} & 83.85\% & 76.53\%\\
			HON4D: Histogram of Oriented 4D Normals for Activity Recognition from Depth Sequences \cite{6618942}& 78.53\%  & 74.03\%  \\
			3-D human action recognition by shape analysis of motion trajectories on riemannian manifold \cite{6894548} & 79.61\% &  62.00\% \\
			\hline
		\end{tabular}
	}
	\caption{Comparison of the accuracy measure among different state-of-the-art approaches on the SHREC dataset.}
	\label{comparison}
\end{table*}

\subsection{Comparisons}
\label{compstate}

We compared the proposed method with key works of the current state-of-the-art presented in~\cite{7789643,6894548,6618942,6595915,desmedt} on the SHREC dataset \cite{desmedt}. The SHREC dataset has been selected as: (a) it provides different types of data to allow comparisons between methods based on different acquisition sensors; (b) it allows the classification of hand gestures with different degrees of complexity; (c) it provides data that allow to extract all the features necessary for the proposed method. The SHREC dataset is also an excellent example of a semaphoric gesture to show the effectiveness of the proposed method on other categories of gesture. The SHREC dataset contains 14 dynamic gestures performed by 28 participants (all participants are right handed) and captured by the Intel RealSense short range depth camera. Each gesture is performed between 1 and 10 times by each participant in two way: using one finger and the whole hand. Therefore, the dataset is composed by 2800 sequences captured. The depth image, with a resolution of 640x480, and the coordinates of 22 joints (both in the 2D depth image space and in the 3D world space) are saved for each frame of each sequence in the dataset. For the proposed method we only needed the 3D coordinates of the joints from which we derived the features of our interest. The depth images and hand skeletons were captured at 30 frames per second and the length of sample gestures ranges from 20 to 170 frames. Since some sequences of the dataset are very short, in order to avoid sampling with a very low $T$ value, we used the padding technique to increase the length of these sequences to an acceptable value of $T$ (i.e., $T=100$). As shown in Table~\ref{comparison}, the proposed method outperforms the accuracy values of the other works, both in the dataset divided into 14 classes and in the dataset divided into 28 classes. The confusion matrices obtained from the tests are shown in Fig.~\ref{comparison_confusion_matrix}.
By analyzing these matrices well we can see that the method can classify very well the gestures performed using only one finger, while in the version using the whole hand some mismatches occur. In details, the gesture 16 (SWIPE LEFT) is sometimes erroneously classified as gesture 26 (SWIPE V) and gesture 8 (PINCH), instead, the gesture 18 (SWIPE UP) is confused with the gesture 6 (EXPAND). By carefully analyzing the variations of the feature values, we notice that the angles obtained from these instances are similar and the movements of the hand in space are not substantial. Despite these isolated cases, the method achieves excellent performance.
This result demonstrates how the DLSTM and the selected features are a very powerful solution in recognizing different types of hand gestures. Although our work is focused on language gestures, the tests carried out on this dataset have highlighted how our method can also handle semaphoric gestures, i.e., gestures that define a set of symbols to communicate with machines.

\section{Conclusion}\label{Conclusion}
In this paper, a novel language hand gesture recognition approach based on a DLSTM is presented. A set of new discriminative features based on both joint angles and fingertip positions are used in combination with DLSTM for the first time to obtain high accuracy in the hand gesture recognition. In addition, a novel dataset based on a large subset of the ASL is created to train and test the proposed method. Moreover, this dataset is used to analyze the effectiveness of the extracted features and the behavior of the network by varying the number of stacked LSTMs. As a future development, we would like to create a public data set based on the ASL composed of a higher number of gestures. A possible improvement of the method could be obtained by integrating the RGB information, as well as the skeleton, in order to solve ambiguous cases (such as those between gesture 6 and gesture W).

\section{Acknowledge}
This work was supported in part by the MIUR under grant "Dipartimenti di eccellenza 2018-2022" of the Department of Computer Science of Sapienza University.

% Can use something like this to put references on a page
% by themselves when using endfloat and the captionsoff option.
\ifCLASSOPTIONcaptionsoff
  \newpage
\fi

% trigger a \newpage just before the given reference
% number - used to balance the columns on the last page
% adjust value as needed - may need to be readjusted if
% the document is modified later
%\IEEEtriggeratref{8}
% The "triggered" command can be changed if desired:
%\IEEEtriggercmd{\enlargethispage{-5in}}

% references section

% can use a bibliography generated by BibTeX as a .bbl file
% BibTeX documentation can be easily obtained at:
% http://mirror.ctan.org/biblio/bibtex/contrib/doc/
% The IEEEtran BibTeX style support page is at:
% http://www.michaelshell.org/tex/ieeetran/bibtex/
\bibliographystyle{IEEEtran}
% argument is your BibTeX string definitions and bibliography database(s)
\bibliography{bare_jrnl}
%
% <OR> manually copy in the resultant .bbl file
% set second argument of \begin to the number of references
% (used to reserve space for the reference number labels box)

% biography section
% 
% If you have an EPS/PDF photo (graphicx package needed) extra braces are
% needed around the contents of the optional argument to biography to prevent
% the LaTeX parser from getting confused when it sees the complicated
% \includegraphics command within an optional argument. (You could create
% your own custom macro containing the \includegraphics command to make things
% simpler here.)
%\begin{IEEEbiography}[{\includegraphics[width=1in,height=1.25in,clip,keepaspectratio]{mshell}}]{Michael Shell}
% or if you just want to reserve a space for a photo:

% insert where needed to balance the two columns on the last page with
% biographies
%\newpage

% You can push biographies down or up by placing
% a \vfill before or after them. The appropriate
% use of \vfill depends on what kind of text is
% on the last page and whether or not the columns
% are being equalized.

%\vfill

% Can be used to pull up biographies so that the bottom of the last one
% is flush with the other column.
%\enlargethispage{-5in}

% that's all folks
\end{document}